\documentclass[conference]{IEEEtran}

\usepackage{cite}
\usepackage{amsmath,amssymb,amsfonts}
\usepackage{graphicx}
\usepackage{textcomp}
\usepackage[table]{xcolor}
\usepackage{booktabs}
\usepackage{tabularx}
\usepackage{array}
\usepackage{pgfplots}

\usepgfplotslibrary{groupplots}
\pgfplotsset{compat=1.18}

\newcolumntype{Y}{>{\centering\arraybackslash}X}

\definecolor{PlotSwin}{RGB}{214,128,153}
\definecolor{PlotReplay}{RGB}{91,169,219}
\definecolor{PlotBase}{RGB}{135,140,148}
\definecolor{PlotWorking}{RGB}{235,174,73}
\definecolor{PlotEpisodic}{RGB}{96,174,92}
\definecolor{PlotOracle}{RGB}{139,100,190}
\definecolor{PlotWrong}{RGB}{190,190,195}

\newcommand{\swinone}[1]{\cellcolor[RGB]{252,232,237}#1}
\newcommand{\swintwo}[1]{\cellcolor[RGB]{246,207,217}#1}
\newcommand{\swinthree}[1]{\cellcolor[RGB]{238,181,196}#1}
\newcommand{\swinfour}[1]{\cellcolor[RGB]{228,156,176}#1}
\newcommand{\swinbest}[1]{\cellcolor[RGB]{214,128,153}\textbf{#1}}

\newcommand{\blueone}[1]{\cellcolor[RGB]{226,242,252}#1}

\newcommand{\bluefour}[1]{\cellcolor[RGB]{126,191,230}#1}
\newcommand{\bluebest}[1]{\cellcolor[RGB]{91,169,219}\textbf{#1}}

\newcommand{\greenone}[1]{\cellcolor[RGB]{230,246,229}#1}
\newcommand{\greentwo}[1]{\cellcolor[RGB]{203,234,199}#1}

\newcommand{\greenfour}[1]{\cellcolor[RGB]{135,197,127}#1}
\newcommand{\greenbest}[1]{\cellcolor[RGB]{96,174,92}\textbf{#1}}

\newcommand{\deitone}[1]{\cellcolor[RGB]{230,246,229}#1}

\newcommand{\deitfour}[1]{\cellcolor[RGB]{135,197,127}#1}
\newcommand{\deitbest}[1]{\cellcolor[RGB]{96,174,92}\textbf{#1}}

\newcommand{\vitone}[1]{\cellcolor[RGB]{226,242,252}#1}

\newcommand{\vitfour}[1]{\cellcolor[RGB]{126,191,230}#1}
\newcommand{\vitbest}[1]{\cellcolor[RGB]{91,169,219}\textbf{#1}}

\newcommand{\routingcell}[1]{%
\cellcolor[RGB]{220,204,239}\textbf{#1}}

\newcommand{\neutralcell}[1]{%
\cellcolor[RGB]{244,244,244}#1}

\newcommand{\effone}[1]{\cellcolor[RGB]{239,244,248}#1}

\newcommand{\effthree}[1]{\cellcolor[RGB]{172,205,225}#1}
\newcommand{\effbest}[1]{\cellcolor[RGB]{121,178,211}\textbf{#1}}

\begin{document}

\title{Where Should Experience Live? Hierarchical Hebbian Memory
for Continual Vision Transformers}

% Replace with the conference-required anonymous format if needed.
\author{
\IEEEauthorblockN{
Mohammed Yusuf Mujawar, 
Noorbakhsh Amiri Golilarz\
}
\IEEEauthorblockA{
Department of Computer Science\\
The University of Alabama, Tuscaloosa, AL, USA\\
}
}

\maketitle

% =========================================================
% ABSTRACT
% =========================================================

\begin{abstract}
Vision Transformers provide strong visual representations but
typically rely on slowly updated parameters, limiting their ability
to organize newly acquired information across different memory
timescales. This work proposes \textit{Hierarchical Hebbian
Memory}, a three-level memory architecture composed of rapid
Working Memory, persistent Routed Episodic Memory, and slower
Semantic Memory. A learned controller regulates memory
contribution, read and write routing, plasticity, retention, and
consolidation. A causal read-before-write lifecycle ensures that
the current outcome cannot influence the prediction it supervises.
The architecture is evaluated on Omniglot 5-way 1-shot
recognition and CORe50 continual object recognition. With
Swin-Tiny, the hierarchical model reaches 97.39\% accuracy on
Omniglot and 95.37\% final accuracy on CORe50 when combined
with experience replay. Learned multi-bank retrieval reaches
47.50\% delayed-association accuracy, compared with 24.17\%
for a single persistent bank and 25.00\% without memory.
After intervening distractors, Episodic Memory retains
approximately 0.96 cosine similarity with stored associations,
while Working Memory falls to approximately 0.05. These
results show that Hebbian association and learned memory
routing can jointly organize online visual experience across
rapid, persistent, and consolidated memory timescales within
Vision Transformers.
\end{abstract}


% =========================================================
% INTRODUCTION
% =========================================================

\section{Introduction}

Vision Transformers provide strong visual representations, but
most of their knowledge is stored in parameters updated through
gradient based training. This limits their ability to rapidly store
new associations without modifying the backbone. The problem is
especially relevant in few shot and continual learning, where useful
information may need to be acquired from only a small number of
observations. Hebbian learning offers a local mechanism for
forming such associations \cite{hebb1949organization}, while fast
weight models maintain rapidly changing memory states alongside
slower learned parameters \cite{hinton1987fast,ba2016fast}.
Associative fast weight formulations further connect key value
binding with Transformer computation
\cite{schlag2021linear,munkhdalai2018metalearning}.

Earlier Hebbian Vision Transformer studies established two useful
results. Fixed Hebbian memory showed that its effectiveness
depends strongly on where associative binding is introduced into
the backbone \cite{money2026where}. Adaptive Hebbian Routing introduced learned control over memory contribution,
plasticity, and retention for each few shot episode
\cite{mujawar2026adaptivehebbianmemoryrouting}. These
approaches improve control of temporary associative memory, but
they do not address how experience should be organized when
information must persist beyond a single episode.

This creates a broader memory allocation problem. A single
persistent state can accumulate experience but may increase
interference when unrelated associations share the same storage.
Conversely, resetting memory after every short context prevents
useful information from being retained. Human memory provides
a useful functional motivation for separating these roles.
Working memory supports short lived information relevant to
current processing \cite{baddeley2003working}, episodic memory
preserves experience together with contextual information
\cite{eichenbaum2017prefrontal,moscovitch2016episodic}, and
semantic memory represents more stable knowledge
\cite{kumar2021semantic}. Consolidation provides a mechanism
through which selected experience can become more persistent
over time \cite{squire2015consolidation}. These principles
motivate a computational hierarchy with different memory
lifetimes.

This work proposes \textit{Hierarchical Hebbian Memory}, which
organizes online visual experience across three interacting levels:
\textit{Hebbian Working Memory}, \textit{Routed Episodic
Memory}, and \textit{Semantic Memory}. Working Memory forms
rapid context specific associations, Episodic Memory maintains
multiple persistent banks with learned read and write routing,
and Semantic Memory selectively consolidates useful episodic
information over a slower timescale. A learned controller
coordinates memory contribution, routing, plasticity, retention,
and consolidation, so that Hebbian plasticity determines what
association is strengthened while routing determines where it is
stored. The architecture follows a causal online lifecycle in
which existing memory is read before prediction and outcome
conditioned updates are applied only afterward. In the primary
continual setting, the controller receives no task identity,
although experience boundaries remain known to the protocol.
Unlike recent memory augmented Transformer systems that focus
mainly on long context language processing or continual
associative learning
\cite{he2025hmt,fountas2025episodic,nechesov2025calm,
omidi2025review}, this work studies how online visual experience
can be distributed across multiple memory timescales using
Hebbian associative learning within Vision Transformers.

The architecture is evaluated in two complementary regimes:
Omniglot 5-way 1-shot recognition for rapid adaptation and
CORe50 continual object recognition across repeated visual
experiences \cite{lomonaco2017core50}. Continual comparisons
include conventional replay and a Latent Replay implementation
adapted from Pellegrini \textit{et al.}
\cite{pellegrini2020latent}. Beyond recognition accuracy, the
evaluation examines delayed retrieval, routing behavior, bank
interventions, memory ablations, semantic consolidation, and
computational cost.

The main contributions are:
\begin{itemize}
    \item A three-level Hierarchical Hebbian Memory architecture
    for Vision Transformers, separating rapid Working Memory,
    persistent Routed Episodic Memory, and slower Semantic
    Memory.

    \item A task ID free controller that regulates both memory
    allocation and memory strength through separate read and
    write routing, plasticity, retention, contribution, and
    consolidation decisions.

    \item An extension of temporary Hebbian associative memory
    into persistent multi-timescale storage using a causal online
    update process.

    \item Evaluation across few shot and continual visual
    learning, together with direct analysis of delayed retrieval,
    routing organization, memory timescales, consolidation, and
    computational efficiency.
\end{itemize}

The remainder of this paper is organized as follows.
Section~II reviews related work on Hebbian and fast weight
memory, neurocognitive memory organization, and
memory-augmented Transformers. Section~III presents the
proposed Hierarchical Hebbian Memory architecture and its
three memory levels. Section~IV describes the experimental
setup and evaluation protocols. Section~V reports the main
results and memory analyses. Section~VI discusses the findings,
and Section~VII concludes the paper.

% =========================================================
% RELATED WORK
% =========================================================

\section{Related Work}

\subsection{Hebbian and Fast Weight Memory}

Hebbian learning provides a local mechanism for strengthening
associations between coactive representations
\cite{hebb1949organization}. Fast weight models extend this
principle by maintaining rapidly changing states alongside slower
parameters learned through gradient based optimization
\cite{hinton1987fast,ba2016fast}. Related approaches have used
temporary associative states for meta learning and differentiable
plasticity \cite{munkhdalai2018metalearning,
miconi2018differentiable}. Fast weight programming further
established a close connection between key value association and
Transformer computation \cite{schlag2021linear}, while a
memory based analysis of Transformers showed how learned
weight matrices can represent associations between embeddings
\cite{bietti2023birth}.

Hebbian fast weights have also been studied directly in Vision
Transformers. Fixed Hebbian memory showed that effectiveness
depends strongly on where associative binding is introduced into
the backbone \cite{money2026where}. Adaptive Hebbian Routing introduced learned control over memory contribution,
plasticity, and retention according to the current few shot episode
\cite{mujawar2026adaptivehebbianmemoryrouting}. These
approaches primarily operate with temporary episode specific
memory. The proposed method considers the broader problem of organizing associative information across persistent episodic storage and slower consolidation..

Increasing persistence changes the learning problem. A temporary
memory state can be reset before unrelated episodes interact,
whereas persistent memory must determine how new observations
should coexist with previously stored associations. The proposed
hierarchy therefore treats association strength and memory
allocation as complementary operations. Hebbian plasticity
determines what association is strengthened, while routing
determines where that association is stored and later retrieved.

\subsection{Neurocognitive Memory Organization}

Human memory is commonly described through interacting
systems with different functions and timescales. Working memory
supports temporary maintenance and manipulation of information
relevant to current processing \cite{baddeley2003working}.
Episodic memory preserves experiences together with contextual
information, whereas semantic memory represents knowledge
that is less dependent on a specific event
\cite{eichenbaum2017prefrontal,moscovitch2016episodic,
kumar2021semantic}. Broader cognitive neuroscience accounts
similarly emphasize that memory is distributed across several
interacting processes rather than represented by a single uniform
store \cite{sridhar2023cognitive,camina2017basis}.

Prefrontal and hippocampal interactions also illustrate how
memory storage can be coordinated with context dependent
encoding and retrieval \cite{eichenbaum2017prefrontal}. This
functional separation motivates the controller and episodic
banks used here, which abstract context dependent storage and
retrieval into computational operations.

Memory consolidation provides another useful principle for
organizing information across timescales. Initially formed
memories can change as experience is reorganized and integrated
into more stable representations
\cite{squire2015consolidation,moscovitch2016episodic}.
Semantic knowledge can consequently become less dependent on
the individual event from which it originated
\cite{kumar2021semantic}. Semantic Memory follows this
functional interpretation by receiving selected information from
persistent Episodic Memory rather than storing every observation
directly. More generally, this approach is consistent with
intelligence inspired by neurocognitive principles, where
biological memory and control mechanisms motivate computational
functions \cite{golilarz2026nii}.

\subsection{Memory Augmented and Continual Transformers}

Recent Transformer systems have explored persistent, episodic,
and hierarchical memory beyond conventional attention. Hierarchical Memory Transformer (HMT)
organizes stored information hierarchically and retrieves relevant
memory for efficient processing of long language contexts
\cite{he2025hmt}. Episodic Memory Language Model (EM LLM) introduces episodic segmentation and
retrieval inspired by properties of human episodic memory
\cite{fountas2025episodic}. These systems separate immediate
processing from information retained across longer contexts,
although their primary objective is long context language
modeling rather than online visual learning.

Associative and persistent storage have also been investigated
through other memory architectures. Continual Associative Learning Model (CALM) studies continual
associative learning using sparse distributed memory
\cite{nechesov2025calm}. A broader review of memory augmented
Transformers describes state based, external, parameter based,
and hybrid memory mechanisms and identifies retrieval,
interference, forgetting, and coordination as recurring challenges
\cite{omidi2025review}.

The proposed method focuses on combining rapid Hebbian
association, learned allocation across several persistent episodic
banks, and slower semantic consolidation within Vision
Transformer learning. Separate read and write routes allow the
location used to store an experience to be controlled independently
from the memory selected for later retrieval.

Continual learning provides a practical setting in which this
organization can be tested. CORe50 was introduced for continuous
object recognition under repeated observations and changing
acquisition sessions \cite{lomonaco2017core50}. Replay reduces
forgetting by presenting earlier information during later
learning. Latent Replay instead stores intermediate
representations that can be reused during later training
\cite{pellegrini2020latent}. In the present evaluation, replay is
an external rehearsal mechanism rather than another level of
the internal memory hierarchy.

% =========================================================
% METHOD
% =========================================================

\section{Hierarchical Hebbian Memory}

\subsection{Architecture Overview}

The proposed architecture organizes online visual experience
across three memory timescales: \textit{Hebbian Working
Memory}, \textit{Routed Episodic Memory}, and \textit{Semantic
Memory}. Working Memory provides rapid temporary association,
Episodic Memory preserves experience through multiple
persistent banks, and Semantic Memory consolidates selected
information on a slower timescale. A learned controller regulates
memory contribution, read and write routing, plasticity,
retention, and consolidation.

\begin{figure*}[!t]
    \centering
    \includegraphics[width=1\textwidth]
    {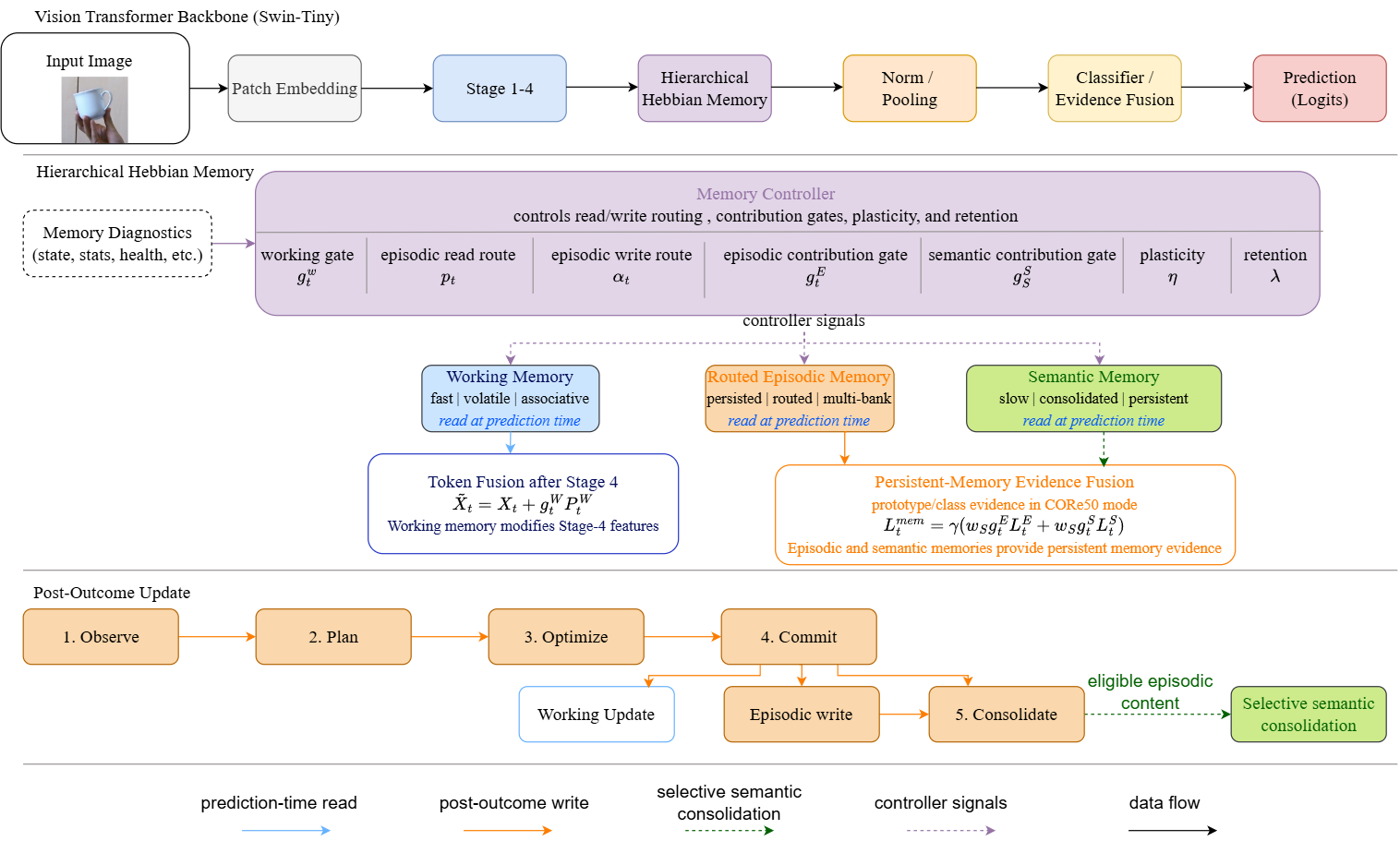}
    \caption{Overview of the proposed Hierarchical Hebbian Memory
    architecture and its memory lifecycle using Swin-Tiny.}
    \label{fig:hierarchical_architecture}
\end{figure*}

Let $X_t\in\mathbb{R}^{T\times d}$ denote the Transformer
representation at time $t$, where $T$ is the number of tokens
and $d$ is the feature dimension. Prediction first uses only the
memory state that existed before the current observation is
written. The outcome can influence memory only after the
prediction has been produced. The complete continual lifecycle
follows the sequence
\textit{read, predict, observe, plan, optimize, commit, and
consolidate}, as illustrated in
Fig.~\ref{fig:hierarchical_architecture}.

The plan stage determines the outcome conditioned memory
operation without immediately changing persistent state.
Parameter optimization is then performed, after which the
detached memory operation is committed. Consolidation occurs
periodically when eligible episodic information is available.
This ordering prevents the current target from entering memory
before the prediction that it supervises.

\subsection{Hebbian Working Memory}

The associative mechanism uses a Hebbian fast-weight formulation related to those employed in prior Vision Transformer memory models
\cite{money2026where,
mujawar2026adaptivehebbianmemoryrouting}. Given $X_t$,
learned projections produce

\begin{equation}
Q_t=X_tW_Q,\qquad
K_t=X_tW_K,\qquad
V_t=X_tW_V
\label{eq:qkv}
\end{equation}

The head index is omitted for clarity. Key and value
coactivation forms the association

\begin{equation}
A_t=
\operatorname{clip}
\left(
\frac{K_t^{\top}V_t}{\sqrt{T}}
\right)
\label{eq:association}
\end{equation}

The resulting association is incorporated into memory through the
general Hebbian update

\begin{equation}
M_t=
\operatorname{clipnorm}
\left(
\lambda_t M_{t-1}
+
\eta_t A_t,c_M
\right),
\label{eq:general_memory}
\end{equation}

where $M_{t-1}$ is the memory state available before the current
write, $\eta_t$ controls the strength of the new association,
$\lambda_t$ controls retention of the previous state, and $c_M$
bounds the memory norm. This general update follows the adaptive
Hebbian formulation
\cite{mujawar2026adaptivehebbianmemoryrouting} and is
specialized below for the different memory levels.

At prediction time, Working Memory is read from the state that
existed before the current observation is written:

\begin{equation}
R_t^{W}=Q_tW_{t-1}
\label{eq:working_read}
\end{equation}

A learned gate regulates the contribution of $R_t^{W}$ to the
backbone representation. After the prediction and observation,
the Working Memory update is

\begin{equation}
W_t=
\operatorname{clipnorm}
\left(
\lambda_t^{W}W_{t-1}
+
\eta_t^{W}A_t,c_W
\right),
\label{eq:working}
\end{equation}

where $\eta_t^{W}$ controls plasticity and $\lambda_t^{W}$
controls retention.

Repeated key and value coactivation reinforces the corresponding
association, following the Hebbian principle
\cite{hebb1949organization}. Working Memory is reset at natural
few-shot episode boundaries and therefore provides rapid
adaptation without carrying temporary associations indefinitely.

\subsection{Semantic Memory and Consolidation}

Semantic Memory provides the slowest level of the hierarchy and
receives selected information from Episodic Memory. Let $C_t$
denote an eligible episodic candidate and $g_t^{S}$ its
consolidation strength. The update is represented as

\begin{equation}
S_t=
\operatorname{Consolidate}
\left(
S_{t-1},g_t^{S}C_t
\right)
\label{eq:semantic}
\end{equation}

Consolidation eligibility depends on accumulated evidence such
as sufficient new writes, memory utility, support within
episodic storage, and consistency across stored information.
Eligible associations are promoted toward the slower Semantic
Memory state, while other information remains episodic. This
introduces a longer retention timescale inspired by memory
consolidation
\cite{squire2015consolidation,moscovitch2016episodic}.

For few-shot learning, the hierarchy uses a
\textit{consolidate then freeze} protocol. Semantic Memory
develops during training and remains fixed during held-out
evaluation, while Working and Episodic Memory are reset for
each evaluation episode.

In continual learning, Semantic Memory persists across
experiences and receives periodic eligible consolidations from
episodic storage. This separates rapidly changing contextual
associations from information retained over longer timescales.

\subsection{Controller and Transformer Integration}

The controller coordinates access to the three memory levels.
It receives the current representation together with memory
state information and produces separate episodic read and write
routes, memory contribution gates, plasticity, and retention
signals. Where configured, the controller also provides
consolidation-related control.

Let $\mathbf{m}_{t-1}$ summarize the memory state available
before the current write. The controller combines this state with
the current pooled representation,

\begin{equation}
\mathbf{c}_t=
\mathcal{C}_{\theta}
\left(
\operatorname{Pool}(X_t),\mathbf{m}_{t-1}
\right),
\label{eq:controller}
\end{equation}

where $\mathcal{C}_{\theta}$ is the learned controller and
$\mathbf{c}_t$ contains the read and write routing decisions,
memory contribution gates, plasticity, and retention signals.
Thus, memory control depends jointly on the current visual
representation and the state of the existing memory hierarchy.

Adaptive Hebbian Routing uses learned contribution, plasticity, and retention for episode-specific fast-weight memory\cite{mujawar2026adaptivehebbianmemoryrouting}. The proposed hierarchical controller instead incorporates these forms of control within persistent memory allocation and retrieval.

During continual learning, existing Working, Episodic, and
Semantic states are read before the current outcome is available.
The model produces its prediction, observes the outcome, plans
the memory operation, and updates gradient trained parameters.
The detached persistent memory operation is then committed.
Persistent memory states are detached between stream steps,
avoiding backpropagation through the complete stream history.

The primary continual controller receives no task identity.
Experience boundaries remain known to the experimental
protocol but are not supplied as routing labels. The resulting
protocol is therefore \textit{task ID free continual learning with
known experience boundaries}.

For Swin-Tiny, memory operates at Stage 4, motivated by reported evidence that late Hebbian binding can be more suitable than dense memory insertion throughout the backbone
\cite{money2026where}. For DeiT-Small and ViT-Small, one
shared hierarchy is read at blocks 3, 6, 9, and 12, with the
persistent write performed at block 12.

% =========================================================
% EXPERIMENTAL SETUP
% =========================================================

\section{Experimental Setup}

\subsection{Datasets and Learning Protocols}

The evaluation considers two complementary settings: few shot
character recognition and continual object recognition. Omniglot
is used for rapid adaptation \cite{lake2015human}, while
CORe50 evaluates persistent memory across a changing visual
stream \cite{lomonaco2017core50}.

For few shot learning, Omniglot is evaluated under a 5-way
1-shot episodic protocol. Each episode contains five classes
with one labeled support example per class. Query samples are
classified using class prototypes formed from support
representations, following Prototypical Networks
\cite{snell2017prototypical}. In this the support samples can
update episode specific memory, whereas query samples are read
only.

Continual learning is evaluated on CORe50 in the New Instances
setting \cite{lomonaco2017core50}. The stream contains repeated
observations of the same object classes across sequential
experiences and changing acquisition conditions. A fixed
complete test set is evaluated after each experience. Episodic
and Semantic Memory persist across the stream, while Working
Memory operates on its shorter timescale.

The controller receives no task identity. Experience boundaries
remain known to the data and evaluation protocol, giving a
\textit{task ID free continual learning setting with known
experience boundaries}. Current labels affect memory only after
the corresponding prediction has been produced, following the
lifecycle in Section~III.

\subsection{Backbones and Comparison Models}

Swin-Tiny is the primary backbone for the main comparison and
detailed memory analysis. The hierarchy is integrated at Stage 4.
DeiT-Small and Vit-Small provides a cross-architecture evaluation using one
shared hierarchy with reads at blocks 3, 6, 9, and 12 and a
persistent write at block 12. The corresponding Transformer
backbones follow their original formulations
\cite{liu2021swin,touvron2021deit,dosovitskiy2021vit}.

For continual learning, the primary Swin comparison includes a
base model without hierarchy or replay, Experience Replay,
Latent Replay, and the full hierarchy combined with matched
Experience Replay. Latent Replay stores intermediate
representations and is implemented as an adaptation of
Pellegrini \textit{et al.} \cite{pellegrini2020latent}. The same
replay capacity is retained for the matched Replay and
Hierarchy + Replay comparisons.

Additional Swin controls isolate individual persistent memory
functions. A single persistent bank tests persistence without
multi-bank routing. A routed Episodic configuration retains
multiple banks while disabling Semantic Memory. The complete
configuration combines Working Memory, routed Episodic Memory,
Semantic Memory, and replay. Detailed mechanism analysis is
centered on Swin, while DeiT and ViT provide architecture
transfer comparisons.

\subsection{Memory and Evaluation Controls}

Few shot and continual experiments use distinct memory
lifecycles. During held-out few shot evaluation, Semantic Memory
retains the state acquired during training but remains frozen. A
semantic state fingerprint is checked before and after evaluation
to verify that the stored state is unchanged. Working and
Episodic Memory are reset for each held-out episode.

In continual learning, Episodic and Semantic Memory persist
across experiences. Each sample first reads the memory state
available at prediction time and produces a prediction. The
outcome is observed afterward. The corresponding write is then
planned, model parameters are optimized, and the persistent
memory operation is committed. Persistent states are detached
between stream steps.

Memory analysis includes frozen memory ablations, separate read
and write routing logs, delayed association probes, and selective
bank interventions. These evaluations examine whether stored
associations remain retrievable and whether particular memory
states have functional effects.

\subsection{Metrics and Implementation}

Few shot performance is measured using held-out episodic
classification accuracy. Continual recognition is evaluated using
final fixed test accuracy and the area under the fixed test
accuracy trajectory across experiences. Memory behavior is
examined through delayed retrieval, routing association with
class or context, episodic bank utilization, routing entropy,
inter-bank similarity, memory ablations, and semantic
consolidation activity.

Efficiency is measured using parameter count, prediction
latency, prediction followed by observation latency, and memory
usage. Prediction timing is separated from the complete
prediction and observation path so that the additional cost of
the online memory operation can be measured directly.

% =========================================================
% RESULTS
% =========================================================

\section{Results and Analysis}

\subsection{Few-Shot Recognition}

The Omniglot experiments first examine whether the proposed
hierarchy can extend temporary Hebbian adaptation across
additional memory timescales while retaining strong few shot
recognition. Table~\ref{tab:omniglot_main} presents the main
Swin-Tiny progression.

\begin{table}[!t]
\caption{Omniglot 5-way 1-shot Swin-Tiny comparison.}
\label{tab:omniglot_main}
\centering
\setlength{\tabcolsep}{3.5pt}
\renewcommand{\arraystretch}{1.08}
\begin{tabularx}{\columnwidth}{@{}lYY@{}}
\toprule
Model & Accuracy (\%) & Params (M) \\
\midrule
Swin Base
& \swinone{96.14}
& \effbest{27.51} \\
Fixed Hebbian
& \swintwo{96.74}
& \effthree{29.87} \\
Adaptive Hebbian
& \swinthree{96.94}
& \effthree{29.94} \\
Hierarchical Hebbian
& \swinbest{97.39}
& \effone{33.56} \\
\bottomrule
\end{tabularx}
\end{table}

Hierarchical Hebbian Memory reaches \textbf{97.39\%}, compared
with 96.14\% for the Swin base, 96.74\% for the fixed Hebbian
configuration, and 96.94\% for Fully Adaptive Hebbian Routing
under the corresponding Swin comparison
\cite{mujawar2026adaptivehebbianmemoryrouting}. Relative to
these reported values, the hierarchy improves the base by
1.25 percentage points, the fixed configuration by 0.65 points,
and Adaptive Hebbian Routing by 0.45 points.

The hierarchical configuration contains 33.56M parameters,
compared with 27.51M for the base backbone and 29.94M for
Adaptive Hebbian Routing. The additional capacity supports the
controller and the three memory levels.

\subsection{Few-Shot Memory Ablations}

Table~\ref{tab:omniglot_ablation} evaluates several memory
allocation, access, and semantic lifecycle controls. All reported controls remain above 97\% accuracy. K=1
Persistent reaches 97.45\%, while learned Top-2 routing reaches
97.36\%. The read-only control reaches 97.49\%, and Semantic
Freeze K=1 reaches 97.60\%. The results show that strong few
shot recognition is maintained across several memory allocation
and lifecycle choices.

\begin{table}[!t]
\caption{Omniglot 5-way 1-shot memory ablations.}
\label{tab:omniglot_ablation}
\centering
\setlength{\tabcolsep}{4pt}
\renewcommand{\arraystretch}{1.05}
\begin{tabularx}{\columnwidth}{@{}lY@{}}
\toprule
Configuration & Accuracy (\%) \\
\midrule
\multicolumn{2}{c}{\textit{Routing and Access Controls}} \\
\midrule
K=1 Persistent & \swinfour{97.45} \\
K=4 Uniform Routing & \swinone{97.12} \\
K=4 Learned Top-2 & \swinthree{97.36} \\
K=4 Random Write & \swinthree{97.36} \\
K=4 Read Only & \swinbest{97.49} \\
K=4 Write Only & \swintwo{97.33} \\
\midrule
\multicolumn{2}{c}{\textit{Semantic Lifecycle Controls}} \\
\midrule
Semantic Freeze K=1 & \swinbest{97.60} \\
Semantic Freeze K=4 Routed & \swintwo{97.27} \\
Semantic Freeze K=4 Uniform & \swinthree{97.30} \\
No Consolidation & \swinone{97.11} \\
\bottomrule
\end{tabularx}
\end{table}

\subsection{Continual Object Recognition}

CORe50 evaluates persistent memory across repeated visual
experiences. Table~\ref{tab:core50_main} reports the highest
completed final fixed test accuracy for each Swin-Tiny
configuration.

\begin{table}[!t]
\caption{CORe50 continual recognition with Swin-Tiny.}
\label{tab:core50_main}
\centering
\setlength{\tabcolsep}{3.2pt}
\renewcommand{\arraystretch}{1.08}
\begin{tabularx}{\columnwidth}{@{}lYY@{}}
\toprule
Configuration & Accuracy (\%) & Params (M) \\
\midrule
Base
& \swinthree{94.75}
& \effbest{27.56} \\
Experience Replay
& \swinbest{95.37}
& \effbest{27.56} \\
Latent Replay
& \swinone{87.17}
& \effbest{27.56} \\
Full Hierarchy + Replay
& \swinbest{95.37}
& \effone{33.62} \\
\bottomrule
\end{tabularx}
\end{table}

Experience Replay and Full Hierarchy + Replay both reach
\textbf{95.37\%} at the displayed precision. The hierarchical
configuration achieves the same displayed peak recognition
accuracy while additionally maintaining Working Memory, eight
routed persistent Episodic Memory banks, and Semantic Memory.

Table~\ref{tab:core50_runs} shows the individual completed Swin results. In Seed 0, Experience Replay and Full Hierarchy + Replay both
display as 95.37\%. The hierarchy obtains the higher final
accuracy in Seeds 1 and 2.

\begin{table}[!t]
\caption{CORe50 final fixed test accuracy (\%) for Swin-Tiny evaluations.}
\label{tab:core50_runs}
\centering
\setlength{\tabcolsep}{2.8pt}
\renewcommand{\arraystretch}{1.06}
\begin{tabularx}{\columnwidth}{@{}lYYY@{}}
\toprule
Configuration & Seed 0 & Seed 1 & Seed 2 \\
\midrule
Base
& \blueone{93.36}
& \greenone{93.22}
& \swinthree{94.75} \\
Experience Replay
& \bluebest{95.37}
& \greentwo{93.75}
& \swintwo{93.60} \\
Full Hierarchy + Replay
& \bluebest{95.37}
& \greenbest{94.19}
& \swinbest{94.82} \\
\bottomrule
\end{tabularx}
\end{table}

\subsection{Online Continual Performance}

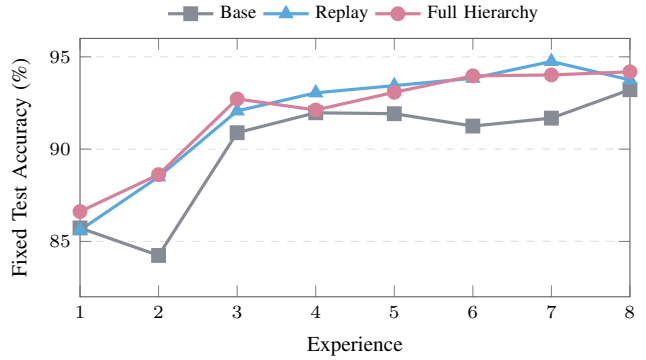
\begin{figure}[!t]
\centering
\begin{tikzpicture}
\begin{axis}[
    width=\columnwidth,
    height=5.0cm,
    xmin=1,
    xmax=8,
    ymin=82,
    ymax=96,
    xtick={1,2,3,4,5,6,7,8},
    xlabel={Experience},
    ylabel={Fixed Test Accuracy (\%)},
    xlabel style={font=\footnotesize},
    ylabel style={font=\footnotesize},
    tick label style={font=\scriptsize},
    legend style={
        font=\scriptsize,
        draw=none,
        fill=none,
        at={(0.5,1.02)},
        anchor=south,
        legend columns=3
    },
    axis line style={black!65},
    major grid style={black!12,dashed},
    ymajorgrids=true,
    mark size=2.2pt
]

\addplot[
    very thick,
    color=PlotBase,
    mark=square*,
    mark options={fill=PlotBase}
] coordinates {
    (1,85.73)
    (2,84.24)
    (3,90.89)
    (4,91.97)
    (5,91.92)
    (6,91.25)
    (7,91.68)
    (8,93.22)
};

\addplot[
    very thick,
    color=PlotReplay,
    mark=triangle*,
    mark options={fill=PlotReplay}
] coordinates {
    (1,85.61)
    (2,88.50)
    (3,92.07)
    (4,93.05)
    (5,93.44)
    (6,93.84)
    (7,94.75)
    (8,93.75)
};

\addplot[
    very thick,
    color=PlotSwin,
    mark=*,
    mark options={fill=PlotSwin}
] coordinates {
    (1,86.62)
    (2,88.62)
    (3,92.72)
    (4,92.12)
    (5,93.08)
    (6,93.97)
    (7,94.02)
    (8,94.19)
};

\legend{Base, Replay, Full Hierarchy}

\end{axis}
\end{tikzpicture}
\caption{CORe50 fixed test accuracy across eight experiences
for Swin-Tiny.}
\label{fig:core50_curve}
\end{figure}

Final accuracy summarizes the end of the continual stream,
whereas the accuracy trajectory shows how recognition develops
across experiences. For the selected Swin-Tiny Seed 1
evaluation shown in Fig.~\ref{fig:core50_curve}, the Base model,
Experience Replay, and Full Hierarchy + Replay reach online
AUC values of 0.9776, 0.9809, and 0.9811, respectively.

\subsection{Cross-Backbone Evaluation}

A cross-backbone evaluation is presented in Table~\ref{tab:core50_backbones} for Swin-Tiny, DeiT-Small and ViT-Small. Swin-Tiny increases from 95.37\% for Replay and Full Hierarchy + Replay in the selected evaluation.
DeiT-Small reaches 93.02\% with Replay and 92.21\% with the
hierarchy. ViT-Small reaches 93.63\% with the hierarchy,
compared with 92.66\% for Replay. The same memory framework
therefore operates across both hierarchical and flat Vision
Transformer backbones, although its effect on recognition varies
with the backbone.

\begin{table}[!t]
\caption{CORe50 cross-backbone comparison.}
\label{tab:core50_backbones}
\centering
\setlength{\tabcolsep}{1.9pt}
\renewcommand{\arraystretch}{1.07}
\begin{tabularx}{\columnwidth}{@{}llYY@{}}
\toprule
Backbone & Configuration & Accuracy & Params \\
& & (\%) & (M) \\
\midrule

Swin-Tiny
& Replay
& \swinbest{95.37}
& \swinbest{27.56} \\

& Full Hierarchy
& \swinbest{95.37}
& \swinone{33.62} \\

\midrule

DeiT-Small
& Replay
& \deitbest{93.02}
& \deitbest{22.57} \\

& Full Hierarchy
& \deitfour{92.21}
& \deitone{24.09} \\

\midrule

ViT-Small
& Replay
& \vitfour{92.66}
& \vitbest{22.57} \\

& Full Hierarchy
& \vitbest{93.63}
& \vitone{24.09} \\

\bottomrule
\end{tabularx}
\end{table}

\subsection{Continual Memory Ablations}

Table~\ref{tab:core50_arch_ablation} compares persistent memory configurations. All configurations remain above 95\% final accuracy
in the comparison. K=1 Persistent reaches the highest
endpoint at 95.55\%, while K=8 Uniform Routing reaches the
highest online AUC at 0.9806. Full Hierarchy + Replay reaches
95.37\% while retaining all three memory levels. The
classification results therefore show that several persistent
memory organizations can maintain strong continual recognition.

\begin{table}[!t]
\caption{Swin-Tiny CORe50 memory ablations.}
\label{tab:core50_arch_ablation}
\centering
\setlength{\tabcolsep}{1.7pt}
\renewcommand{\arraystretch}{1.05}
\begin{tabularx}{\columnwidth}{@{}lYYY@{}}
\toprule
Configuration & Accuracy & AUC & Params \\
& (\%) & & (M) \\
\midrule
Base
& \swinone{93.36}
& \greenone{0.9774}
& \bluebest{27.56} \\
Replay
& \swinfour{95.37}
& \greentwo{0.9803}
& \bluebest{27.56} \\
K=1 Persistent + Replay
& \swinbest{95.55}
& \greentwo{0.9803}
& \bluefour{33.61} \\
K=8 Uniform + Replay
& \swinthree{95.17}
& \greenbest{0.9806}
& \blueone{33.62} \\
K=8 Routed Episodic + Replay
& \swintwo{95.08}
& \greenfour{0.9805}
& \blueone{33.62} \\
Full Hierarchy + Replay
& \swinfour{95.37}
& \greenfour{0.9805}
& \blueone{33.62} \\
\bottomrule
\end{tabularx}
\end{table}

\subsection{Delayed Retrieval and Memory Timescales}

A real-image delayed association evaluation tests whether stored
information remains accessible after intervening observations.
The learned multi-bank condition corresponds to the proposed
routing mechanism, where the controller selects the episodic
storage and retrieval route without receiving the correct context
identity.

Learned Multi-Bank retrieval reaches \textbf{47.50\%}, compared
with 24.17\% for a Single Persistent Bank and 25.00\% without
memory. Forcing retrieval through an incorrect bank reduces
accuracy to 0.00\%. These results show that successful delayed
recall depends not only on storing an association, but also on
retrieving it from an appropriate persistent memory state.

\begin{figure*}[!t]
    \centering
    \includegraphics[width=0.7\textwidth]{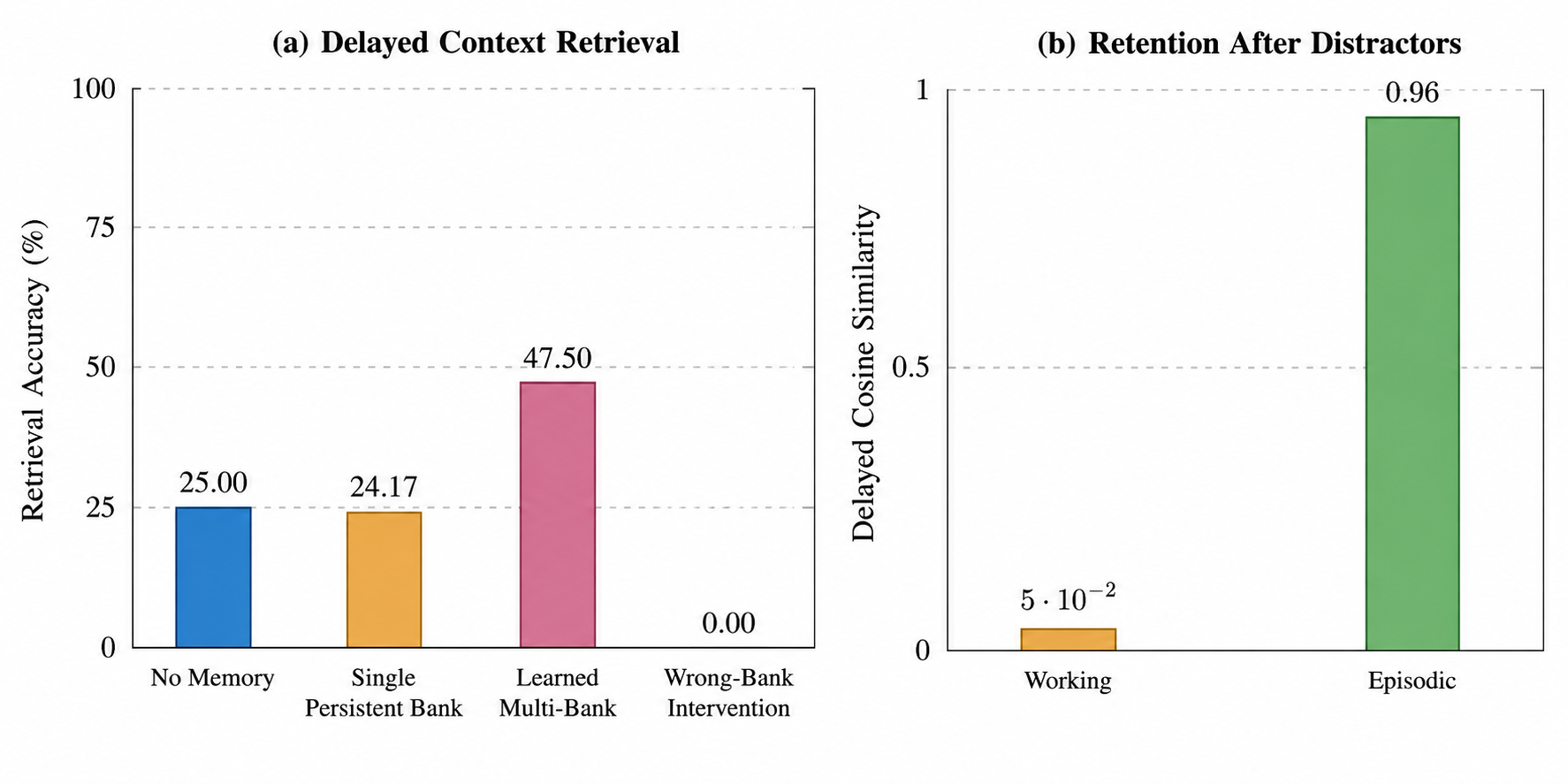}
    \caption{Functional memory evaluation on CORe50.
    (a) Delayed context retrieval under learned and intervention
    conditions.
    (b) Memory retention after intervening distractors.}
    \label{fig:memory_mechanisms}
\end{figure*}

Fig.~\ref{fig:memory_mechanisms} summarizes the learned and
intervention-based retrieval conditions together with the
retention difference between Working and Episodic Memory.

Learned Multi-Bank routing improves delayed retrieval by
23.33 percentage points over the Single Persistent Bank and by
22.50 points over the No Memory condition. The Wrong-Bank
Intervention further shows that the persistent storage location
has a direct functional effect on later retrieval.

A separate oracle control supplies the known correct context bank
during storage and retrieval. Under this condition, retrieval
reaches \textbf{100.00\%}. The oracle provides an
upper-bound control showing that the stored association is fully
recoverable when the correct episodic bank is known.

The retention experiment provides complementary evidence for
the intended memory timescales. After intervening distractors,
Episodic Memory retains approximately \textbf{0.96} cosine
similarity with the stored association, whereas Working Memory
falls to approximately \textbf{0.05}. Across the completed
full-hierarchy evaluations, episodic delayed similarity remains
approximately 0.97, 0.97, and 0.96. This separation supports the
intended role of Working Memory as a rapidly changing
associative state and Episodic Memory as a longer-lived
persistent store.

\subsection{Routing Organization and Semantic Consolidation}

\begin{table}[!t]
\caption{Episodic routing organization in the Swin-Tiny
hierarchy.}
\label{tab:routing_analysis}
\centering
\setlength{\tabcolsep}{4pt}
\renewcommand{\arraystretch}{1.08}
\begin{tabularx}{\columnwidth}{@{}lY@{}}
\toprule
Routing Measure & Result \\
\midrule
\neutralcell{Active Episodic Banks}
& \routingcell{8 / 8} \\
\neutralcell{Class-Routing NMI}
& \routingcell{0.56} \\
\neutralcell{Normalized Utilization Entropy}
& \routingcell{0.98} \\
\neutralcell{Mean Off-Diagonal Bank Cosine}
& \routingcell{0.62} \\
\bottomrule
\end{tabularx}
\end{table}

Table~\ref{tab:routing_analysis} show all eight episodic banks remain active during continual
learning. A class-routing NMI of 0.56 indicates a relationship
between object class and learned bank allocation. Normalized
utilization entropy of 0.98 shows that routing remains broadly
distributed across the available banks rather than collapsing
onto a small subset. The mean off-diagonal bank cosine of 0.62
shows that the persistent bank states are neither identical nor
completely independent within the shared representation space.

The controlled delayed association evaluation provides an
additional routing measurement, with context-routing NMI of
approximately 0.49. Together with the bank-sensitive retrieval
results, this indicates that learned routing influences where
context dependent associations remain accessible.

Semantic Memory updates on a slower schedule. Across the
primary CORe50 hierarchy evaluations, the semantic store accepts
4, 3, and 5 consolidations while rejecting 372, 409, and 408
candidate promotions. During few shot training, Semantic Memory
also undergoes consolidation, while its fingerprint remains
unchanged throughout held-out evaluation. These results verify a
selective slower update lifecycle rather than direct promotion of
every episodic write.

\subsection{Efficiency}

The efficiency benchmark illustrated in Table~\ref{tab:swin_efficiency} Swin with different configurations and a batch size of 128. Full Hierarchy increases the Swin-Tiny model from 27.56M to
33.62M parameters. Prediction latency increases from 55.84 ms
to 61.27 ms. For the hierarchy, prediction followed by
observation and the associated memory operation requires
67.41 ms, corresponding to approximately 6.14 ms beyond
prediction alone. Peak allocated benchmark VRAM increases from
2.40 GB to 2.58 GB.

\begin{table}[!t]
\caption{Swin-Tiny CORe50 efficiency.}
\label{tab:swin_efficiency}
\centering
\setlength{\tabcolsep}{1.5pt}
\renewcommand{\arraystretch}{1.06}
\begin{tabularx}{\columnwidth}{@{}lYYYY@{}}
\toprule
Model &
Params &
\shortstack{Predict\\Time} &
\shortstack{Predict+\\Observe} &
\shortstack{Peak\\VRAM} \\
&
(M) &
(ms) &
(ms) &
(GB) \\
\midrule
Replay
& \bluebest{27.56}
& \greenbest{55.84}
& \swinbest{55.98}
& \routingcell{2.40} \\
Full Hierarchy
& \blueone{33.62}
& \greenone{61.27}
& \swinone{67.41}
& \cellcolor[RGB]{240,232,248}2.58 \\
\bottomrule
\end{tabularx}
\end{table}

% =========================================================
% DISCUSSION
% =========================================================

\section{Discussion}

The results show that Hierarchical Hebbian Memory can extend
rapid associative adaptation into persistent multi-timescale
memory while maintaining strong visual recognition. On
Omniglot, the hierarchical Swin configuration reaches
97.39\%, compared with 96.14\% for the Swin base, 96.74\%
for fixed Hebbian memory, and 96.94\% for Adaptive Hebbian
Routing as reported in\cite{mujawar2026adaptivehebbianmemoryrouting}. The few shot ablations
also remain above 97\% across multiple routing, access, and
semantic lifecycle controls.

CORe50 provides a complementary view under continual visual
experience. Experience Replay and Full Hierarchy + Replay both
reach 95.37\% for Swin. The hierarchy therefore
maintains the same displayed peak recognition accuracy while
simultaneously operating Working Memory, multiple persistent
Episodic Memory banks, and Semantic Memory.

The delayed association experiment provides a direct test of the
functional value of routed persistent storage. Learned multi-bank
retrieval reaches 47.50\%, compared with 24.17\% for a single
persistent bank and 25.00\% without memory. Forcing retrieval
through an incorrect bank reduces accuracy to 0\%. These
results show that delayed recall depends both on storing an
association and on retrieving an appropriate persistent state.
The 100\% oracle result further confirms that the underlying
association remains recoverable when its correct storage
location is supplied.

The hierarchy also exhibits a clear separation between memory
timescales. Episodic Memory retains approximately 0.96 cosine
similarity after distractors, whereas Working Memory falls to
approximately 0.05. This behavior matches their intended
functions: Working Memory provides rapidly changing local
association, while Episodic Memory retains information over
longer delays. Routing measurements complement this result.
All eight episodic banks remain active, class-routing NMI reaches
0.56, and utilization entropy of 0.98 shows broad use of the
available storage. Semantic Memory introduces the slowest update timescale.
Only selected episodic candidates are consolidated during
CORe50, while the few shot evaluation verifies that the learned
semantic state remains unchanged during held-out evaluation.
Together, the three levels separate rapid association, persistent
contextual storage, and slower consolidation within a common
memory architecture. Future work can investigate boundary-free streams, richer consolidation
criteria, adaptive memory capacity, and larger scale continual
visual settings.

% =========================================================
% CONCLUSION
% =========================================================

\section{Conclusion}

This work introduced Hierarchical Hebbian Memory for Vision
Transformers, extending rapid associative learning into three
interacting memory timescales: Working Memory, Routed
Episodic Memory, and Semantic Memory. A learned controller
coordinates memory contribution, read and write routing,
plasticity, retention, and consolidation, while a causal
read-before-write lifecycle prevents current outcomes from
influencing the predictions they supervise. The hierarchical Swin configuration reaches 97.39\% on
Omniglot and 95.37\% final accuracy on CORe50 with Experience
Replay. Mechanism evaluations further show that learned multi-bank retrieval reaches 47.50\%, compared
with 24.17\% for a single persistent bank and 25.00\% without
memory. Episodic Memory also retains approximately 0.96 cosine
similarity after intervening distractors, while Working Memory
falls to approximately 0.05. Together, these results show that Hebbian association and learned
memory allocation can be combined within Vision Transformers
to organize online visual experience across rapid, persistent,
and consolidated memory timescales.

\section*{Acknowledgment}

The authors acknowledge the support and resources provided by the
Bioinspired Robotics, AI, Imaging and Neurocognitive Systems (BRAINS)
Laboratory at The University of Alabama.

\bibliographystyle{IEEEtran}
\bibliography{references_final}

\end{document}